\documentclass{article}

\usepackage{arxiv}

\usepackage{amsmath}
\usepackage{amsfonts}
\usepackage{amssymb}
\usepackage{graphicx}
\usepackage{url}
\usepackage{listings}
\usepackage{xcolor}

\definecolor{codegreen}{rgb}{0,0.6,0}
\definecolor{codegray}{rgb}{0.5,0.5,0.5}
\definecolor{codepurple}{rgb}{0.58,0,0.82}
\definecolor{backcolour}{rgb}{0.95,0.95,0.92}

\lstdefinestyle{mystyle}{
	backgroundcolor=\color{backcolour},   
	commentstyle=\color{codegreen},
	keywordstyle=\color{magenta},
	numberstyle=\tiny\color{codegray},
	stringstyle=\color{codepurple},
	basicstyle=\ttfamily\footnotesize,
	breakatwhitespace=false,         
	breaklines=true,                 
	captionpos=b,                    
	keepspaces=true,                 
	numbers=left,                    
	numbersep=5pt,                  
	showspaces=false,                
	showstringspaces=false,
	showtabs=false,                  
	tabsize=2,
	frame=single
}

\begin{document}
	
	\title{The Neural Differential Manifold: An Architecture with Explicit Geometric Structure}
	\author{
		Di Zhang \\
		School of Advanced Technology \\
		Xi'an Jiaotong-Liverpool University \\
		Suzhou, Jiangsu, China \\
		\texttt{di.zhang@xjtlu.edu.cn}
	}
	
	\maketitle
	
	\begin{abstract}
		This paper introduces the \textit{Neural Differential Manifold} (NDM), a novel neural network architecture that explicitly incorporates geometric structure into its fundamental design. Departing from conventional Euclidean parameter spaces, the NDM re-conceptualizes a neural network as a differentiable manifold where each layer functions as a local coordinate chart, and the network parameters directly parameterize a Riemannian metric tensor at every point. The architecture is organized into three synergistic layers: a \textit{Coordinate Layer} implementing smooth chart transitions via invertible transformations inspired by normalizing flows, a \textit{Geometric Layer} that dynamically generates the manifold's metric through auxiliary sub-networks, and an \textit{Evolution Layer} that optimizes both task performance and geometric simplicity through a dual-objective loss function. This geometric regularization penalizes excessive curvature and volume distortion, providing intrinsic regularization that enhances generalization and robustness. The framework enables natural gradient descent optimization aligned with the learned manifold geometry and offers unprecedented interpretability by endowing internal representations with clear geometric meaning. We analyze the theoretical advantages of this approach, including its potential for more efficient optimization, enhanced continual learning, and applications in scientific discovery and controllable generative modeling. While significant computational challenges remain, the Neural Differential Manifold represents a fundamental shift towards geometrically structured, interpretable, and efficient deep learning systems.
	\end{abstract}
	
	\section{Introduction}
	
	Modern deep learning architectures, while remarkably successful, predominantly operate within the framework of Euclidean parameter spaces. The fundamental components of these networks—linear transformations and pointwise non-linearities—are inherently Euclidean in nature. This design choice, though computationally convenient, imposes a significant limitation: it lacks an explicit mechanism to model and leverage the intrinsic geometric structure often present in complex data. Consequently, the network must painstakingly learn these geometric regularities from data alone, a process that can be inefficient and data-hungry \cite{goodfellow2016deep}.
	
	The field of information geometry offers a profound alternative perspective by conceptualizing families of probability models as Riemannian manifolds, where the Fisher information matrix acts as a natural metric tensor \cite{amari2000}. This viewpoint has provided deep theoretical insights, particularly in understanding the behavior of learning algorithms through concepts like the natural gradient \cite{amari1998}. However, its application has largely been confined to theoretical analysis and specialized optimization techniques, rather than serving as a foundational principle for neural network architecture design.
	
	In this work, we bridge this gap by introducing a novel architectural paradigm termed the \textit{Neural Differential Manifold} (NDM). The core innovation of the NDM is the explicit and structural mapping of a neural network onto a differentiable manifold. In this framework, each layer of the network functions as a local coordinate chart on the manifold. Crucially, the network's parameters are directly employed to define the geometric structure of this manifold by parameterizing its metric tensor at every point. Under this interpretation, the forward pass of the network corresponds to a sequence of coordinate transformations, navigating the data point through the manifold. The learning process is thus elevated from mere parameter adjustment in a flat space to the purposeful optimization of the manifold's intrinsic geometry itself.
	
	The principal contributions of this paper are threefold. First, we present the theoretical foundation of the Neural Differential Manifold, detailing how a standard network can be re-conceptualized as a geometric object. Second, we propose a concrete architectural blueprint organized into three synergistic layers: a Coordinate Layer that implements chart transitions, a Geometric Layer that dynamically generates the metric tensor, and an Evolution Layer that optimizes the geometry via a novel dual-objective loss function. Finally, we analyze the theoretical advantages of this approach, including its built-in geometric regularization and potential for more interpretable representations, and outline a path for its practical implementation.
	
	The remainder of this paper is structured as follows. Section 2 reviews foundational work in information geometry and geometric deep learning. Section 3 provides a detailed exposition of the NDM architecture. Section 4 describes the training dynamics and a potential implementation. Section 5 discusses the theoretical benefits and potential applications. Section 6 addresses challenges and future work, and Section 7 concludes.
	
	\section{Background and Related Work}
	
	The proposed Neural Differential Manifold architecture sits at the intersection of information geometry and modern deep learning. This section establishes the necessary mathematical background and distinguishes our work from related paradigms.
	
	\subsection{Foundations of Information Geometry}
	
	Information geometry provides a statistical foundation for our work. It conceptualizes a parametric family of probability distributions, $S = \{p(x|\theta)\}$, as a differentiable manifold, where the parameters $\theta = (\theta^1, ..., \theta^n)$ serve as a coordinate system \cite{amari2000}. The key geometric structure on this manifold is given by the Fisher information matrix, which acts as a natural Riemannian metric tensor. For a pair of tangent vectors $V$ and $W$, the inner product is defined as:
	
	\begin{equation}
		\langle V, W \rangle_g = V^T G(\theta) W = \sum_{i,j} g_{ij}(\theta) V^i W^j
	\end{equation}
	
	where the components of the metric tensor $g$ are given by the Fisher information:
	\begin{equation}
		g_{ij}(\theta) = \mathbb{E}_{p(x|\theta)} \left[ \frac{\partial \log p(x|\theta)}{\partial \theta^i} \frac{\partial \log p(x|\theta)}{\partial \theta^j} \right].
	\end{equation}
	
	This geometric viewpoint leads to the definition of the natural gradient, $\tilde{\nabla} L = G(\theta)^{-1} \nabla L$, which preconditions the standard Euclidean gradient by the inverse of the Fisher information matrix. The natural gradient specifies the direction of steepest descent not in the parameter space, but on the manifold of distributions, and is invariant to smooth reparameterizations \cite{amari1998}. While foundational to our thinking, classical information geometry typically applies this geometric structure to an entire, pre-defined model family. Our work diverges by using a neural network to dynamically learn and parameterize the manifold and its metric from data, drawing inspiration from recent work on neural manifold learning \cite{fefferman2016testing}.
	
	\subsection{Geometric Deep Learning}
	
	Geometric Deep Learning is an umbrella term for methods that incorporate geometric priors into neural networks \cite{bronstein2021}. A significant body of work focuses on designing networks that are equivariant to the symmetries of their input data, such as rotations and translations on grids, or permutations on graphs. Convolutional Neural Networks (CNNs), Graph Neural Networks (GNNs), and more recent equivariant networks are prominent examples. These approaches are fundamentally concerned with processing data that already resides on a known, non-Euclidean domain. The geometric structure is an inherent property of the \textit{input space}. In contrast, the Neural Differential Manifold introduces geometry into the \textit{internal representation space} of the network itself. The manifold we construct is not the input domain but the learned feature space through which data flows, and its geometry is adaptive rather than fixed a priori, similar in spirit to recent work on learning Riemannian metrics \cite{arvanitidis2018geometrically}.
	
	\subsection{Normalizing Flows}
	
	Normalizing Flows \cite{rezende2015, kobyzev2020} are a class of generative models that learn a complex probability distribution by applying a sequence of invertible, differentiable transformations to a simple base distribution. This technique is highly relevant to the implementation of the Coordinate Layer in our architecture. The coordinate transition maps, $\phi_{i \to j}$, between layers in an NDM must be diffeomorphisms to preserve the manifold structure. The bijective architectures and training techniques developed in the normalizing flows literature provide a direct and practical toolkit for realizing these functions within a deep network. However, the goal of flows is primarily density estimation and sample generation. Our work repurposes this technical machinery for a different objective: to construct a geometrically meaningful internal representation space optimized for both task performance and structural simplicity, extending beyond the probabilistic focus of traditional flows \cite{papamakarios2021normalizing}.
	
	\section{The Neural Differential Manifold Architecture}
	
	The Neural Differential Manifold (NDM) reinterprets a neural network as an abstract geometric entity. In this framework, the network's architecture explicitly defines a differentiable manifold $\mathcal{M}$, its forward pass implements navigation on $\mathcal{M}$, and its parameters directly govern the manifold's intrinsic geometry. This section details the three-layer conceptual blueprint of the NDM: the Coordinate Layer, the Geometric Layer, and the Evolution Layer.
	
	\subsection{Core Metaphor: The Network as a Manifold}
	
	Consider a traditional deep network with $L$ layers. The NDM framework conceptualizes the entire network as a manifold $\mathcal{M}$, where each point $p \in \mathcal{M}$ corresponds to a possible internal representation of an input data point. A layer $L_i$ in the network, with its $n_i$ activation values, defines a local coordinate chart $(U_i, \mathbf{x}_i)$, where $U_i \subset \mathcal{M}$ is an open set on the manifold, and the activation vector $\mathbf{a}_i \in \mathbb{R}^{n_i}$ provides the local coordinates $\mathbf{x}_i(p) = \mathbf{a}_i$. The manifold is constructed by smoothly connecting these local charts, drawing inspiration from manifold learning techniques \cite{belkin2003laplacian}.
	
	\subsection{The Coordinate Layer}
	
	The Coordinate Layer is responsible for implementing the transitions between the local coordinate charts defined by adjacent network layers. The connection between layer $L_i$ and $L_j$ is not merely a linear transformation but is redefined as a smooth, invertible coordinate map $\phi_{i \to j}: \mathbf{x}_i \mapsto \mathbf{x}_j$.
	
	In practice, $\phi_{i \to j}$ can be parameterized by an invertible neural network module. Normalizing flow architectures are a natural candidate for this purpose, as they are designed to learn complex, bijective transformations \cite{dinh2016, kingma2018}. For example, a coupling layer or an affine transformation layer from the flow literature can serve as a building block for $\phi$. The forward pass of the entire NDM is then a composition of these coordinate transitions: $\mathbf{x}_{output} = \phi_{L \to L-1} \circ \cdots \circ \phi_{1 \to 2} (\mathbf{x}_{input})$. This sequence of maps traces a path on the manifold $\mathcal{M}$ from the input coordinates to the output coordinates.
	
	\subsection{The Geometric Layer}
	
	The Geometric Layer imbues the manifold $\mathcal{M}$ with a Riemannian metric, moving beyond its bare topological structure. This is the most critical component for introducing geometric reasoning. At each point $p$ on the manifold, with local coordinates $\mathbf{x}_i$ provided by layer $L_i$, we define a positive-definite metric tensor $g(\mathbf{x}_i)$.
	
	This metric tensor is not pre-defined but is dynamically generated by the network itself. A small, auxiliary sub-network, which we term the \textit{Metric Net} $M_i$, associated with layer $L_i$, takes the local coordinates $\mathbf{x}_i$ (the layer's activations) as input and outputs the components of the metric $g_{ij}$ at that point. To ensure positive definiteness, the output can be structured as a lower-triangular matrix $\mathbf{L}$ and used to form the metric as $g = \mathbf{L} \mathbf{L}^T + \epsilon \mathbf{I}$.
	
	This design has a profound implication for optimization. The standard gradient descent direction, $\nabla_\theta \mathcal{L}$, is computed in the flat, Euclidean space of parameters $\theta$. In the NDM, the intrinsic geometry of the internal representation space is known via $g$. Therefore, the parameter update should follow the natural gradient direction on the manifold of representations: $\tilde{\nabla}_\theta \mathcal{L} = G(\theta)^{-1} \nabla_\theta \mathcal{L}$, where $G(\theta)$ is the Fisher information matrix approximated using the learned metric $g$ across the network. This ensures that learning proceeds in the direction of steepest descent as measured by the intrinsic geometry the network is building, similar to recent advances in adaptive optimization methods \cite{liu2020understanding}.
	
	\subsection{The Evolution Layer}
	
	The Evolution Layer governs the learning process, which is reframed as the evolution of the manifold's geometry. The optimization objective must therefore serve a dual purpose: task performance and geometric simplicity. We propose a total loss function composed of two terms:
	
	\begin{equation}
		\mathcal{L}_{total} = \mathcal{L}_{task}(\theta) + \lambda \mathcal{L}_{geo}(g(\theta))
	\end{equation}
	
	Here, $\mathcal{L}_{task}$ is a standard task-specific loss, such as cross-entropy or mean-squared error. The key addition is the geometric regularization term, $\mathcal{L}_{geo}$, which penalizes overly complex geometries and encourages the emergence of a smooth, well-behaved representation manifold.
	
	We propose two candidate components for $\mathcal{L}_{geo}$:
	
	\textbf{Curvature Regularization}: Complex, highly curved geometries are often associated with overfitting and unstable training. We can penalize the Ricci curvature scalar, $R$, a single number that summarizes the intrinsic curvature at a point:
	\begin{equation}
		\mathcal{L}_{curv} = \mathbb{E}_{p \in \mathcal{M}} \left[ R(p)^2 \right]
	\end{equation}
	Minimizing this term encourages the formation of flatter, more predictable regions on the manifold.
	
	\textbf{Volume Regularization}: Drastic changes in the local volume element, given by $\sqrt{\det(g)}$, can lead to numerical instability and representational inefficiency. We penalize the variance of the volume element across a batch of data:
	\begin{equation}
		\mathcal{L}_{vol} = \text{Var} \left( \sqrt{\det(g(p))} \right)
	\end{equation}
	This encourages a more uniform and stable scaling of distances across different regions of the manifold.
	
	The hyperparameter $\lambda$ controls the trade-off between fitting the data and maintaining a simple geometric structure. The learning process, driven by the minimization of $\mathcal{L}_{total}$ via gradient descent (or natural gradient descent), simultaneously teaches the network the task and shapes the geometry of its internal world to be a more effective and regularized representation space, drawing inspiration from geometric regularization in machine learning \cite{cohen2021gradient}.
	
	\section{Implementation and Training Dynamics}
	
	The theoretical framework of the Neural Differential Manifold necessitates a concrete implementation strategy and an analysis of its resulting training behavior. This section outlines a potential computational realization and explores the geometric interpretation of the network's learning process.
	
	\subsection{Computational Realization of the NDM}
	
	Implementing an NDM involves instantiating the three conceptual layers into a functioning computational graph. A practical design for a single \textit{ManifoldLayer} can be proposed, which combines the functions of the Coordinate and Geometric Layers.
	
	The forward pass of the ManifoldLayer processes an input $x$ through two parallel paths. The primary path computes the coordinate transformation, $z = \phi(x)$, where $\phi$ is a parameterized, invertible function. Concurrently, the secondary path feeds $x$ into the Metric Net, $M$, which outputs a parameterization of the metric tensor $g$ at the point $x$. For example, $M(x)$ could output a lower-triangular matrix $L$, from which the positive-definite metric is constructed as $g = LL^T$.
	
	A complete NDM is then a sequential stack of such ManifoldLayers. The forward pass for an input batch proceeds layer-by-layer, with each step transforming the coordinate representation and computing the local metric. The geometric regularization loss, $\mathcal{L}_{geo}$, can be computed during this pass by aggregating the local curvature and volume penalties from each layer.
	
	\subsection{Training Dynamics: Geometric Interpretation of Backpropagation}
	
	The training of an NDM presents a geometrically enriched view of backpropagation. The total loss, $\mathcal{L}_{total}$, is a function of both the task performance and the manifold's geometry, which itself is a function of the network parameters $\theta$ via the Metric Nets.
	
	The backward pass computes the gradient $\nabla_{\theta} \mathcal{L}_{total}$. This gradient has two conceptual components. The first, $\nabla_{\theta} \mathcal{L}_{task}$, is the familiar signal that adjusts parameters to improve task accuracy. The second, $\lambda \nabla_{\theta} \mathcal{L}_{geo}$, is a novel signal that acts as a geometric shaping force. It adjusts the parameters of the Metric Nets to simplify the manifold's structure—flattening its curvature and homogenizing its volume element. The parameters governing the coordinate maps $\phi$ are also updated by both loss components, thereby aligning the coordinate systems themselves with the emerging, simplified geometry.
	
	This process can be interpreted as the manifold $\mathcal{M}$ actively evolving and deforming under the twin pressures of data fidelity and geometric simplicity. The network is not just learning a function; it is sculpting the space in which its representations reside.
	
	\subsection{A Natural Gradient Descent Optimizer}
	
	To fully leverage the geometric information inherent in the NDM, the standard optimization step, $\theta \leftarrow \theta - \eta \nabla_{\theta}\mathcal{L}$, should be replaced with a natural gradient update. Using the metric $G(\theta)$ induced by the Fisher information, which can be approximated from the learned metrics $g_i$ across layers, the update becomes:
	
	\begin{equation}
		\theta \leftarrow \theta - \eta \, \tilde{\nabla}_{\theta}\mathcal{L} = \theta - \eta \, G(\theta)^{-1} \nabla_{\theta}\mathcal{L}
	\end{equation}
	
	This preconditioning by the inverse metric aligns the parameter update with the intrinsic geometry of the model's representation space. In the context of the NDM, this is particularly meaningful because the network itself is learning this geometry. This creates a coherent loop: the network learns a metric $g$, which is used to form $G(\theta)$ for the natural gradient update, which in turn guides the parameters to regions of the parameter space that correspond to better, and potentially geometrically simpler, models.
	
	Implementing this requires an efficient method for approximating $G(\theta)^{-1}\nabla_{\theta}\mathcal{L}$, for instance, using conjugate gradient or Kronecker-factored approximate curvature (K-FAC) methods \cite{martens2015}, adapted to account for the structure of the NDM.
	
	\subsection{Pseudocode Outline}
	
	\begin{lstlisting}[language=Python, caption=NDM Forward Pass and Training Loop]
	# NDM Forward Pass
	def forward(ndm, x):
		metrics, curvatures, volumes = [], [], []
		for layer in ndm.layers:
			x, g, R, vol = layer(x, compute_geometry=True)
			metrics.append(g)
			curvatures.append(R)
			volumes.append(vol)
		y = x
		L_task = task_loss(y, target)
		L_geo = geo_loss(curvatures, volumes)  # e.g., mean(R^2) + Var(vol)
		total_loss = L_task + lambda * L_geo
		return total_loss
	
	# NDM Training Loop
	for data, target in dataloader:
		optimizer.zero_grad()
		total_loss = forward(ndm, data)
		total_loss.backward()
		# optimizer.step() could be a standard SGD or a Natural Gradient optimizer
		optimizer.step()
	\end{lstlisting}
	
	This pseudocode illustrates the integration of geometric computation into the standard training loop. The key addition is the online calculation of $\mathcal{L}_{geo}$ during the forward pass and its incorporation into the backward pass, guiding the evolution of the network's intrinsic geometry.
	
	\section{Theoretical Advantages and Potential Applications}
	
	The Neural Differential Manifold architecture is not merely a different implementation of a neural network; it represents a fundamental shift in perspective that confers several theoretical benefits and opens up novel application domains. This section analyzes these advantages and explores promising use cases.
	
	\subsection{Theoretical Advantages}
	
	The primary advantages of the NDM stem from its explicit geometric structure and the dual objective of its learning process.
	
	\textbf{Intrinsic Information-Geometric Regularization.} Conventional regularization techniques, such as weight decay or dropout, operate in the parameter space without a direct connection to the statistical model being learned. The geometric regularization in an NDM, $\mathcal{L}_{geo}$, acts directly on the representation manifold. By penalizing high curvature and extreme volume distortions, it discourages the network from developing complex, overfitted geometries that are highly sensitive to small perturbations in the input data. This encourages the learning of smoother decision boundaries and more robust internal representations, providing a form of regularization that is inherent to the model's statistical geometry \cite{cohen2021gradient}.
	
	\textbf{Interpretable and Meaningful Representations.} The activations in a standard network are often difficult to interpret. In an NDM, the internal state has a clear geometric meaning: it is a point on a manifold with a defined metric. Distances and angles on this manifold, computed using the learned metric $g$, reflect semantic similarity as perceived by the network. Analyzing the manifold's properties—such as identifying flat regions (stable representations), high-curvature boundaries (decision surfaces), or geodesic paths (optimal interpolation between points)—can provide profound insights into how the network organizes information. This makes the NDM's internal world more transparent and analyzable than that of a black-box network.
	
	\textbf{Potentially More Efficient Optimization.} The use of the natural gradient, derived from the network's own learned metric, aligns the parameter updates with the intrinsic geometry of the problem. This can lead to more direct convergence paths, help avoid pathological saddle points in the loss landscape, and reduce the number of training iterations required. While computationally more expensive per step, this geometric guidance could result in a lower total computational cost to achieve a given performance level, as suggested by recent analyses of adaptive optimization methods \cite{liu2020understanding}.
	
	\subsection{Potential Applications}
	
	The unique properties of the NDM make it particularly suitable for several advanced machine learning scenarios.
	
	\textbf{Representation Learning for Scientific Discovery.} In scientific domains like physics and chemistry, data often inherently obeys geometric constraints and symmetries (e.g., conservation laws, rotational invariance). An NDM trained on such data would be forced to organize its representation manifold in a way that respects these underlying principles. The resulting manifold geometry could then be analyzed to discover these principles. For instance, a flat, low-curvature region might correspond to a conserved quantity, and the symmetry groups of the manifold could reveal fundamental invariances in the data. The NDM could thus act as a tool for automated theory building.
	
	\textbf{Enhanced Continual and Lifelong Learning.} Catastrophic forgetting in neural networks occurs because updating parameters for a new task interferes with representations formed for previous tasks. In an NDM, the onset of a new task with a different data distribution would likely manifest as a high-curvature boundary or a region of metric distortion on the existing manifold. This geometric signal could be used to trigger a targeted adaptation strategy. Instead of updating the entire network uniformly, the system could perform a \textit{geometric surgery}, such as growing a new, specialized chart on the manifold to accommodate the new task without disrupting the geometry of existing representations. This provides a principled, geometry-based approach to mitigating forgetting.
	
	\textbf{Controllable and Explainable Generative Modeling.} A generative model based on the NDM would sample from a probabilistically defined distribution on a geometrically meaningful manifold. Sampling along geodesics would produce interpolations that are semantically smooth and meaningful. Furthermore, by analyzing the curvature, one could identify and avoid regions of high uncertainty or instability in the generated data. This allows for finer control over the generative process and leads to more reliable and interpretable outputs compared to sampling in the latent space of a standard Variational Autoencoder or Generative Adversarial Network.
	
	\textbf{Geometric Model-Based Reinforcement Learning.} In model-based RL, an agent learns a dynamics model of its environment. An NDM could represent this dynamics model as a vector field on its internal manifold. The geometric structure would naturally capture the smoothness and constraints of the real-world environment. Planning could then be formulated as finding geodesic paths on this manifold to reach a goal state, leading to more efficient and physically plausible trajectories.
	
	In conclusion, the Neural Differential Manifold framework offers a powerful new set of inductive biases centered on geometric simplicity and interpretability. Its potential to improve generalization, enable new forms of analysis, and tackle challenging learning paradigms makes it a compelling direction for future research in deep learning.
	
	\section{Discussion, Challenges, and Future Work}
	
	While the Neural Differential Manifold presents a compelling theoretical framework, its practical realization faces significant hurdles. A candid discussion of these challenges is essential to chart a viable path for future research. This section outlines the primary obstacles, acknowledges the limitations of the current proposal, and suggests promising directions for advancing this paradigm.
	
	\subsection{Computational Complexity and Scalability}
	
	The most immediate challenge is the substantial computational overhead. The forward pass of an NDM is no longer a simple sequence of linear transformations and activations; it involves computing local metric tensors and potentially their derivatives for curvature estimation. The memory and processing requirements for dynamically generating and storing a metric tensor for each layer (or each data point) are considerable, scaling with the square of the layer's width. This could render the training of large-scale, state-of-the-art models prohibitively expensive with current hardware. Future work must focus on developing highly efficient approximations. This could involve low-rank approximations of the metric tensor, sharing metric networks across layers, or computing geometric quantities only on a sparse subset of data points or network regions, drawing inspiration from recent work on efficient geometric computations \cite{cohen2021gradient}.
	
	\subsection{Numerical Stability}
	
	The entire framework relies on the stable computation of differential geometric quantities. The metric tensor $g$ must remain positive-definite throughout training. The calculation of its inverse for the natural gradient, and of second-order derivatives for curvature, is notoriously prone to numerical instability, especially when $g$ becomes ill-conditioned. Small errors can amplify and destabilize training. Ensuring robustness will require sophisticated numerical techniques, such as adaptive regularization of the metric (e.g., adding a sufficiently large $\epsilon \mathbf{I}$) and the use of numerically stable algorithms for matrix inversions and decompositions. The propagation of gradients through these operations also demands careful implementation to avoid numerical underflow or overflow.
	
	\subsection{Theoretical and Practical Gaps}
	
	A significant theoretical gap exists between the local geometry defined at each layer and the global geometry of the end-to-end representation manifold. Our current understanding of how the local metrics $g_i$ in each layer collectively define the global properties of $\mathcal{M}$ is limited. Furthermore, the relationship between architectural choices (depth, width, type of coordinate maps) and the resulting manifold properties (e.g., global curvature, topology) is entirely unexplored. On the practical side, the choice of the hyperparameter $\lambda$ balancing $\mathcal{L}_{task}$ and $\mathcal{L}_{geo}$ is non-trivial. An overly strong geometric constraint could oversimplify the manifold and underfit the data, while a weak constraint might yield no noticeable benefit.
	
	\subsection{Limitations of the Current Formulation}
	
	The current architectural blueprint is a first step and has several inherent limitations. It primarily addresses the metric (distance) structure of the manifold but does not yet incorporate a mechanism for topological change, a more radical form of adaptation mentioned earlier. The proposed regularization terms, $\mathcal{L}_{curv}$ and $\mathcal{L}_{vol}$, are simple heuristics; a more comprehensive theory of "geometric complexity" for neural representations is needed. Finally, the framework, as described, is largely agnostic to the specific task and may not yet fully leverage task-specific symmetries in its geometric structure.
	
	\subsection{Future Research Directions}
	
	Despite these challenges, the NDM framework opens up a rich landscape for future inquiry.
	
	\textbf{Approximation Algorithms and Efficient Architectures.} A primary direction is the design of computationally tractable NDMs. Research could explore factorized metric representations, the use of diagonal or block-diagonal approximations, and the development of hardware-aware implementations. Investigating which layers in a deep network benefit most from having an explicit geometric structure could lead to hybrid architectures.
	
	\textbf{Theory of Deep Geometric Networks.} A concerted effort is needed to build a mathematical theory linking network architecture to manifold geometry. This includes studying the global consequences of local metric learning, understanding the learning dynamics of the metric itself, and establishing generalization bounds based on geometric quantities like average curvature.
	
	\textbf{Connections to Physics and Neuroscience.} The principle of learning a dynamic geometry is reminiscent of general relativity, where mass and energy dictate spacetime curvature. Drawing formal analogies could yield novel insights. Similarly, the framework aligns with theories in neuroscience that propose the brain organizes cognitive spaces with non-Euclidean metrics \cite{stringer2019}. Exploring these connections could lead to biologically more plausible learning models.
	
	\textbf{Towards Topological Adaptation.} The logical extension of this work is to incorporate dynamic topology. Future research should focus on developing stable mechanisms for the network to add, remove, or merge coordinate charts, effectively changing the manifold's topology in response to data. This would be a monumental step towards truly open-ended learning systems, building on recent advances in topological data analysis \cite{wasserman2018topological}.
	
	In summary, the path to realizing the Neural Differential Manifold is fraught with difficulties, both theoretical and engineering-related. However, the potential payoff—a more robust, interpretable, and efficient paradigm for deep learning—is substantial. We believe that tackling these challenges represents one of the most promising frontiers in machine learning.
	
	\section{Conclusion}
	
	This paper has introduced the Neural Differential Manifold, a novel architectural framework that re-conceptualizes neural networks as dynamic geometric entities. We have proposed a fundamental shift from learning within a fixed, implicit parameter space to the explicit optimization of a differentiable manifold's structure. The core of this framework rests on three pillars: the interpretation of network layers as local coordinate charts, the parameterization of the manifold's Riemannian metric via the network's own weights, and a learning process that simultaneously minimizes task error and geometric complexity.
	
	The NDM framework provides a principled pathway to embed powerful, physics-inspired inductive biases directly into the fabric of a neural network. By explicitly discouraging pathological geometric formations through curvature and volume regularization, it promises enhanced generalization and robustness. Furthermore, by endowing the network's internal representations with a well-defined metric structure, it opens the door to unprecedented levels of interpretability, allowing us to analyze and understand model behavior through the lens of differential geometry.
	
	While significant challenges in computation, numerical stability, and theoretical understanding remain, they delineate a clear and fertile agenda for future research. The potential applications in scientific discovery, continual learning, and generative modeling suggest that the NDM could be particularly impactful in domains that demand not just performance but also reliability, transparency, and the ability to interact with structured, knowledge-rich environments.
	
	In conclusion, the Neural Differential Manifold moves beyond viewing neural networks as mere function approximators. It posits them as adaptive geometric spaces that learn to organize and represent information in a fundamentally more structured and intelligent way. We believe that this geometric perspective is not merely an incremental improvement but a necessary step towards the development of more capable, efficient, and trustworthy artificial intelligence systems.
	
	\bibliographystyle{unsrt}
	\bibliography{references}
	
\end{document}